\documentclass[runningheads]{llncs}
\usepackage[T1]{fontenc}
\usepackage{graphicx}
\usepackage{tabularx}
\begin{document}
\title{Stage‑Aware Diagnosis of Diabetic Retinopathy via Ordinal Regression}
%
%
\author{Saksham Kumar\inst{1} \and
D Sridhar Aditya\inst{1} \and
T Likhil Kumar\inst{1} \and
Thulasi Bikku\inst{1} \and
Srinivasarao Thota\inst{1} \and
Chandan Kumar\inst{2}\thanks{Corresponding author: chandan.jha150286@gmail.com}}
\authorrunning{Kumar et al.}
\institute{Amrita School of Computing, Amrita Vishwa Vidyapeetham, 
Amaravati Campus, Andhra Pradesh - 522503, India \and
Amity Institute of Information Technology, 
Amity University Jharkhand, Ranchi, India}
\maketitle              
\begin{abstract}
Diabetic Retinopathy (DR) has emerged as a major cause of preventable blindness in recent times. With timely screening and intervention, the condition can be prevented from causing irreversible damage. The work introduces a state-of-the-art Ordinal Regression-based DR Detection framework that uses the APTOS-2019 fundus image dataset. A widely accepted combination of preprocessing methods—Green Channel (GC) Extraction, Noise Masking, and CLAHE—was used to isolate the most relevant features for DR classification. Model performance was evaluated using the Quadratic Weighted Kappa, with a focus on agreement between results and clinical grading. Our Ordinal Regression approach attained a QWK score of 0.8992, setting a new benchmark on the APTOS dataset.

\end{abstract}
\section{Introduction}

\verb|Diabetic Retinopathy|(DR) is amongst the top challenges in ophthalmology-based healthcare. DR is one of the leading causes of preventable blindness globally, amongst a wide range of adults (\cite{Leasher2016Global2010}). It is a microvascular complication of \textbf{Diabetes Mellitus} (commonly known as Type-II Diabetes) which damages the retina's blood vessels, progressively deteriorating vision with a plethora of pathological changes, like vascular leakage, swelling, and abnormal neovascularization. 

DR is detected via a series of stages, each visible through multiple alternations and increasing vision damage. The first level \textit{mild Non-Proliferative Diabetic Retinopathy} is characterised by minuscule swelling in parts of retinal blood vessels and microaneurysms. The condition progresses with the formation of more blood vessels.  New and abnormal blood vessels occupy the aqueous humour, leading to leakage, vision impairment, and complete blindness, which is called 'proliferative diabetic retinopathy' (PDR).

\begin{figure}
    \centering
    \includegraphics[width=1\linewidth]{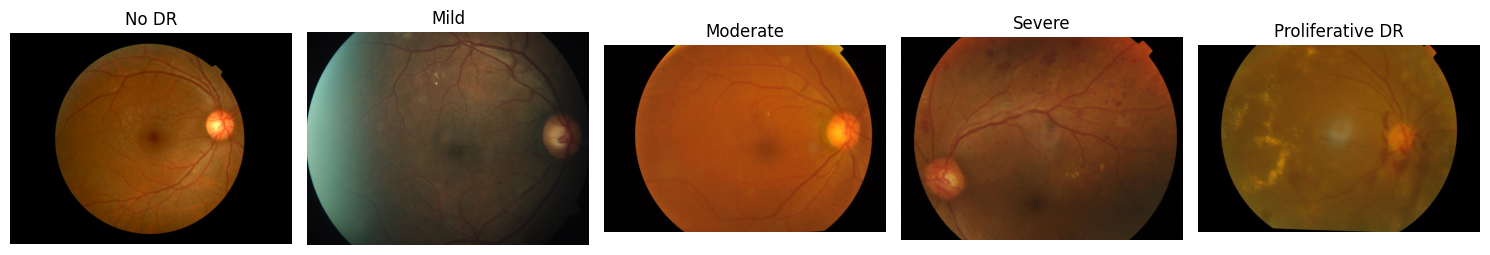}
    \caption{Stages of Diabetic Retinopathy (per ICDRSS)}
    \label{fig:placeholder}
\end{figure}

The Asia Pacific Tele-Ophthalmology Society (APTOS) 2019 Blindness Detection dataset (\cite{Kaggle2019APTOSKaggle}) is a dataset for DR detection of 3,622 fundus photographs. Our primary work is the development of a state-of-the-art automated DR classification model, which labels the fundus amongst five categories of diabetic retinopathy severity. The labels range from no disease (Class 0) to proliferative diabetic retinopathy (Class 4) as per the International Clinical Diabetic Retinopathy Disease Severity Scale  (ICDRSS). 

Recent research emphasises the effectiveness of transfer learning using pre-trained convolutional neural networks for DR classification. Dixit et al.\cite{Dixit2025FundusBlock}  proposed a novel approach utilizing \textit{EfficientNet-B3 with squeeze-and-excitation (SE) blocks}, achieving 88.44\% accuracy on the APTOS-2019 dataset. The integration of SE blocks helped to power the model, focusing on crucial features by implementing channel-wise attention mechanisms. 

The fusion of attention mechanisms is a progressing trend in medical diagnosis, including DR classification. Farag et al.\cite{Farag2022AutomaticModule} combined \textit{DenseNet-169 with Convolutional Block Attention Module (CBAM)}, which enhances discriminative power. 97\% accuracy in binary classification and 82\% accuracy in severity grading on the ICDRSS scale were scored by the model. In summation, the CBAM refocused the model's capability on key features, filtering out distractions.

Lalithadevi et al. \cite{Lalithadevi2024DiabeticTechnique} creared \textit{OptiDex}, by integrating \textit{NASNet-Mobile} with enhanced Cat Swarm optimization and xAI. This approach achieved high accuracy on the test-set and helped in model inference reasoning. It was a major milestone in transparent and explained decision-making in medical applications, in contrast to the DL based black-box predictions.

Bodapati et al.'s \cite{Bodapati2024Self-adaptivePrediction} self-adaptive ensemble for DR achieved an accuracy of 86.22\% and 0.8965 QWK. Their dual-attention model for lesion attention, combined with spatial correlation learning, proved an effective self-adaptive meta learner.

Oulhadj et al. \cite{Oulhadj2022DiabeticRegistration} used deformable registration with a multi-CNN voting ensemble for DR stages grading. The approach worked on aligning the fundus image and background space influence, improving classification accuracy by image standardization. The work got a 0.75 QWK score.



\subsection{Research Gaps}

DR Detection has made significant progress within the multi-label image classification domain. Yet many challenges are yet to be solved,

\begin{enumerate}

    \item Complexity of Feature Extraction: DR involves subtle retinal changes like haemorrhages and microaneurysms. Irrelevant features need to be discarded, with focus on  the pertinent ones. 
    
    \item Imbalanced Class Count: DR Datasets are imbalanced. Normal cases dominate over severe stages, and are under represented. It degrades the classification quality, primarily for the severe stages. 
    
    \item Preprocessing Standardization: Model scoring metrics are highly influenced by the data preprocessing step. Standardized pipelines are needed for consistent results across datasets and imaging conditions.

    \item Standardization of the Kappa Metric within Healthcare: Accuracy alone is insufficient in medical AI. The Kappa metric penalizes prediction errors, hence it is a reliable choice for critical deployments.

\end{enumerate}

The work uses a ResNet-based Ordinal Regression Model to capture the pertinent Fundus features. The study combines a standardized preprocessing pipeline with a robust Deep Learning model. Green Channel Isolation, CLAHE enhancement, median filtering for noise reduction, and standardized resizing with 3-channel recreation for optimal input quality across datasets are used. The work focuses on achieving state-of-the-art performance metrics, using quadratic weighted kappa as the primary evaluation metric. The particular emphasis is put on improving performance on minority classes (severe DR stages).

\section{Methodology}

\begin{figure}
    \centering
    \includegraphics[width=1\linewidth]{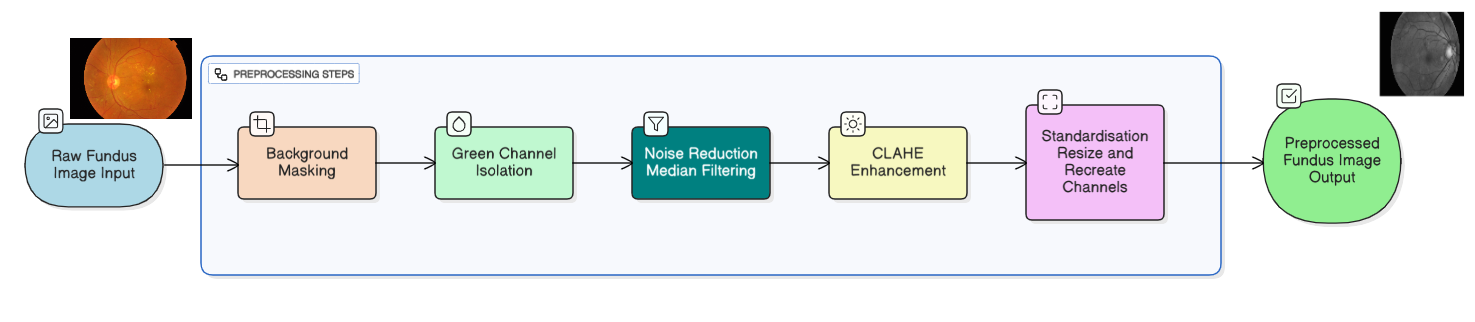}
    \caption{Preprocessing Pipeline}
    \label{fig:Preprocessor}
\end{figure}

For standardizing input images and enhancing relevant features for DR pathology (including lesions, macular edema, hemorrhages, aneurysms, etc), a uniform preprocessing pipeline is applied to every image before it is fed into the models (Figure \ref{fig:Preprocessor}). The pipeline comprises:

\begin{itemize}
    \item \verb|Background Masking|: Removes the black background to isolate the region of interest (ROI).

    \item \verb|Green Channel Isolation|: Extracts the green channel, which best highlights retinal blood vessels and pathological features, improving DR signal clarity. The green channel provides the best contrast for viewing retinal vasculature and lesions, such as microaneurysms and haemorrhages.

    \item \verb|Median Filtering|: A median filter is applied to the isolated green channel, isolating salt-and-pepper noise and other minor artifacts, without blurring the edges of important pathological features.

    \item \verb|CLAHE (Contrast Limited Adaptive Histogram Equalization)|: It divides the whole image into separate parts and enhances contrast in each part individually. It is effective at making micro aneurysms, hard exudates, and other subtle lesions more prominent, helping in effective, clearer grading.

    \item \verb|Standardization by Resizing|: The single-channel enhanced image is replicated across three channels to create a 3-channel grayscale image. The input shape requirements of pre-trained models (ResNet, ViT) are 3-channel, hence it helps to accommodate that requirement. Finally, all images are resized to a standard dimension of 224x224 pixels.
\end{itemize}

\subsection{Model Architecture}

\begin{figure}
    \centering
    \includegraphics[width=1\linewidth]{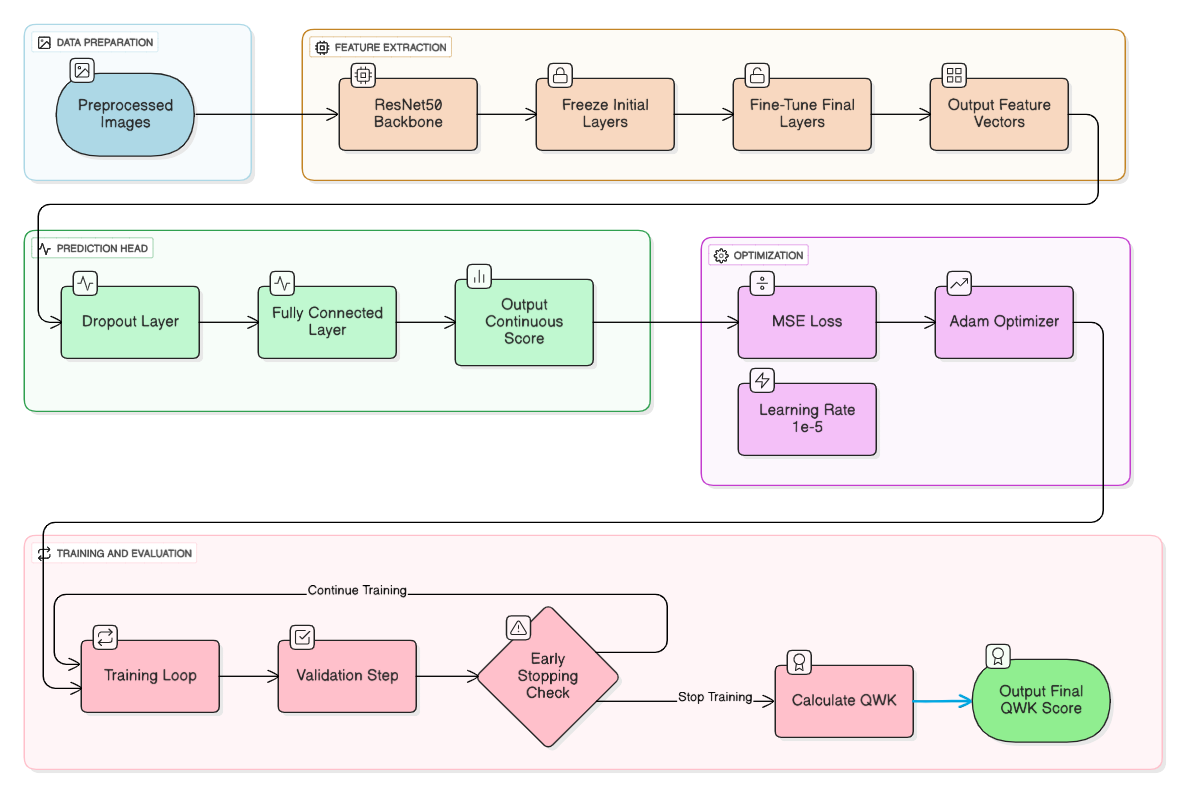}
    \caption{Ordinal Regression with ResNet50 Model Architecture}
    \label{fig:ordreg}
\end{figure}
The methodology treats the problem as an ordinal regression task, using a standard ResNet50 architecture \cite{He2015DeepRecognition} as the backbone (Figure \ref{fig:ordreg}).

A pre-trained ResNet50 model is adapted by replacing its final classification layer with a new head comprising a Dropout layer and a single Dense output neuron.

The model is trained to predict a continuous severity score. A key aspect of the training strategy involves fine-tuning only the final layers of the ResNet50 model while keeping the earlier layers frozen.

Mean Squared Error (MSE) is used to train the model on continuous outputs. For classification, the model's continuous output is clamped to the range [0, 4], rounded to the nearest integer to determine the predicted DR stage.

\section{Results}

Accurate Classification of Diabetic Retinopathy (DR) requires an evaluation metric that reflects the clinical severity of misdiagnosis. Due to the ordinality character of the DR grading scale (stages 0 through 4), standard metrics like \textit{classification accuracy} or the \textit{F1-score} are insufficient, as they see all misclassifications equally. The central metric for this comparative study is the Quadratic Weighted Kappa (QWK). The QWK metric is a score-weighted variation of Cohen's Kappa, which measures inter-rater reliability. It also integrates the weight of disagreement between predicted scores and ground-truth labels \cite{Doewes2023EvaluatingScoring}.

\[ QWK = 1 - \frac{\sum_{i,j} W_{i,j} O_{i,j}}{\sum_{i,j} W_{i,j} E_{i,j}} \]

\begin{figure}
    \centering
    \includegraphics[width=0.6\linewidth]{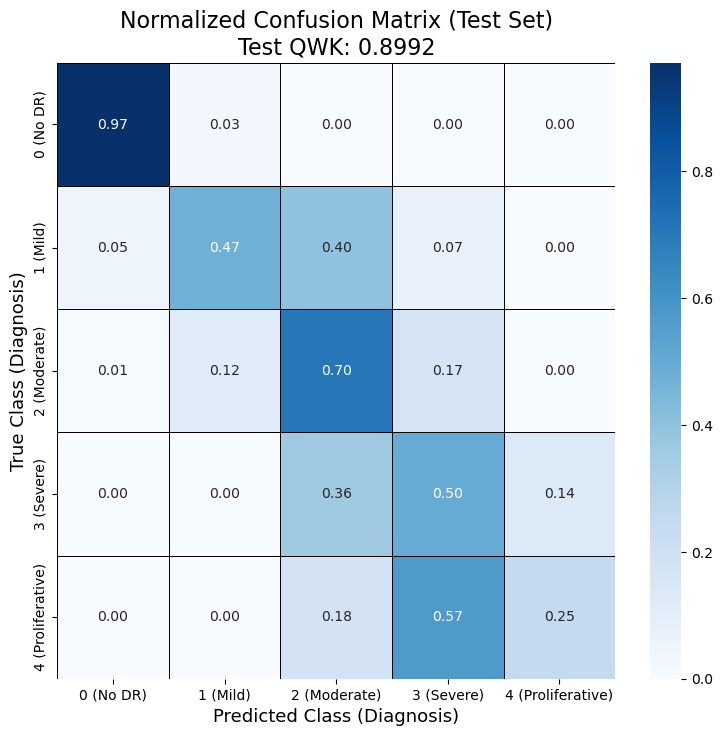}
    \caption{Ordinal Regression Test Set Confusion Matrix}
    \label{fig:ordinalreg_cf}
\end{figure}

The ResNet50-based Ordinal Regression model achieved the highest ever performance, registering a peak Validation Quadratic Weighted Kappa (QWK) of 0.8992, with a Validation Mean Squared Error (MSE) loss of 0.3871 (Figure \ref{fig:ordinalreg_cf}).

\begin{table}[!htbp]
    \centering
    \begin{tabular}{|l|l|r|}\hline
        \textbf{Author} & \textbf{Model} & \textbf{QWK} \\\hline
        Oulhadj et al 2022\cite{Oulhadj2022DiabeticRegistration} & Deformable Registration & 0.75 \\\hline
        Dixit and Jha, 2025\cite{Dixit2025FundusBlock} & Squeeze Excitation Enet & 0.88 \\\hline
        Bodapati and Balaji, 2023\cite{Bodapati2024Self-adaptivePrediction} & Self Stacking Ensemble & 0.89 \\\hline
        \textbf{Proposed Method} & \textbf{Ordinal Regression}* & \textbf{0.8992} \\\hline
    \end{tabular}
    \caption{Comparative Metrics for Proposed Work (*Best Performance)}
    \label{tab:model_stats}
\end{table}

Table \ref{tab:model_stats} clearly illustrates the model's high QWK, compared with recent works in DR detection and classification. The novel Ordinal Regression with a ResNet50-based model performed better. The incorporation of a Fundus Aligned pipeline resulted in a better-performing model on the APTOS 2019 dataset for DR 5-stage classification.

\section{Conclusion and Future Work}

The work developed a distinct machine learning paradigm for the ordinal classification of Diabetic Retinopathy (DR) into five severity grades (0 to 4). A unified preprocessing pipeline — involving Green Channel Isolation, CLAHE, and Median Filtering — was applied throughout the data corpus. 

The performance of the Ordinal Regression with ResNet50 model was primarily evaluated using the Quadratic Weighted Kappa (QWK) metric, which appropriately accounts for the severity of misclassification errors inherent in ordinal grading. The model surpassed the ICDRSS QWK threshold of 0.8.

Through the results were encouraging, several limitations were identified. The GC processor's bulkiness added significant overhead, extending training time. It also fell short to fully capture stage-specific features. More work is required to identify optimal preprocessing methods for Fundoscopy feature extraction. 
The diagnostic transparency of the results can be improved by Explainable AI. Improving interpretability of models such as Ordinal Regression and Hybrid CNN-ViT builds trust in automated grading and critical clinical deployments.


\bibliographystyle{splncs04}
\bibliography{references}
\end{document}